\pdfoutput=1

\documentclass[11pt]{article}

\usepackage{ACL2023}

\usepackage{times}
\usepackage{latexsym}

\usepackage[T1]{fontenc}

\usepackage{graphicx}
\usepackage{float}
\usepackage{color, colortbl}
\usepackage{xcolor}
\definecolor{Gray}{gray}{0.9}
\usepackage{makecell}
\usepackage{lipsum}
\usepackage{float}
\usepackage{verbatim} 
\usepackage{placeins}
\usepackage{hyperref}
\usepackage{epstopdf}
\usepackage{algorithm}
\usepackage{amsmath}
\usepackage{amssymb}
\usepackage{lscape}
\usepackage{pifont}
\usepackage{multirow}
\usepackage{subcaption}
\usepackage{tcolorbox}
\usepackage{graphicx}
\usepackage{supertabular,booktabs}
\usepackage{geometry}
\usepackage{longtable}
\usepackage{pdflscape} 
\usepackage{xcolor}
\usepackage{amsmath}
\usepackage{hyperref}
\usepackage{graphicx}
\usepackage{geometry}
\usepackage{longtable}

\usepackage[utf8]{inputenc}

\usepackage{microtype}
\usepackage{cuted}
\usepackage{inconsolata}

\usepackage{newfloat}

\title{\textbf{\textit{Infogen}}: Generating Complex Statistical Infographics from Documents}

\author{%
  \begin{tabular}[t]{@{}c@{}} 
    Akash Ghosh$^{1}$\thanks{Work done during internship at Adobe Research.}\quad
    Aparna Garimella$^{2}$\quad
    Pritika Ramu$^{2}$\\[2pt]                 
    Sambaran Bandyopadhyay$^{2}$\quad
    Sriparna Saha$^{1}$\\[6pt]                
    $^{1}$ Indian Institute of Technology Patna, India\\
    $^{2}$Adobe Research\\[4pt]
    \small%
    \texttt{\{akash\_2321cs19,sriparna\}@iitp.ac.in}, 
    \texttt{\{garimell,pramu\}@adobe.com} ,
    \texttt{samb.bandyo@gmail.com}
  \end{tabular}
}

\begin{document}

\maketitle
\begin{abstract}

Statistical infographics are powerful tools that simplify complex data into visually engaging and easy-to-understand formats. Despite advancements in AI, particularly with LLMs, existing efforts have been limited to generating simple charts, with no prior work addressing the creation of complex infographics from text-heavy documents that demand a deep understanding of the content.  We address this gap by introducing the task of generating {\it statistical 
infographics} composed of multiple sub-charts (e.g., line, bar, pie) that are contextually accurate, insightful, and visually aligned. 
To achieve this, we define infographic metadata, that includes its title and textual insights, along with sub-chart-specific details such as their corresponding data, alignment, etc. We also present \textbf{\textit{Infodat}}, the first benchmark dataset for text-to-infographic metadata generation, where each sample links a document to its metadata. We propose \textbf{\textit{Infogen}}, a two-stage framework where fine-tuned LLMs first generate metadata, which is then converted into infographic code. Extensive evaluations on \textbf{\textit{Infodat}} demonstrate that  \textbf{\textit{Infogen}} achieves state-of-the-art performance, outperforming both closed and open-source LLMs in text-to-statistical infographic generation. The sample datapoints from \textbf{\textit{Infodat}} can be accessed through this \href{https://github.com/AkashGhosh/Infogen/tree/main}{link}


\end{abstract}

\section{Introduction}
Data visualization tools transform raw data into meaningful visuals, making complex information more accessible and understandable \cite{sadiku2016data}. 
In this work, we focus on \textit{\textbf{Complex statistical infographics}}, which we view as those statistical visuals that are composed of more than one sub-chart, each with its own axes and titles, that are particularly significant for presenting intricate data in the form of simpler components.\footnote{While the term {\it infographic} can refer to a wide range of visual illustrations, we restrict ourselves to only those visuals that only comprise of multiple statistical charts.} 
The common sub-charts include bar, pie, line graphs, and histograms, including their stacked or grouped variants, accompanied by varying textual information in the form of their titles or summaries. 
These information-dense visuals organize complex statistical data into a single cohesive layout, enabling a quick understanding of key statistics and trends.  Widely used in business, academia, and media, they play a crucial role in supporting effective communication and informed decision-making \cite{siricharoen2013infographics}.

\par

With recent advancements in AI, particularly the development of Large Language Models (LLMs) and Visual Language Models (VLMs) \cite{ghosh2024exploring}, the field of data visualization has undergone a transformative evolution \cite{vazquez2024llms}. Generative AI's growing adoption has encouraged users to rely on prompts for diverse tasks \cite{gozalo2023survey}, For example, tools like Adobe Express \cite{adobe_express} and PiktoChart \cite{piktochart} leverage generative AI to analyze textual prompts and propose visually appealing, template-based designs. However, while effective for general infographics, these tools struggle with complex statistical contexts requiring more precise and nuanced data representation. This application of generating complex statistical infographics from text is especially useful for applications such as real-time dashboards from meeting summaries, scientific posters from research abstracts, personalized health visualizations, educational materials, survey summaries, market trends, social media insights, and policy briefs.

\par

\begin{figure*}[hbt!]
    
    \centering
    
    \includegraphics[height=5cm, width=12cm]{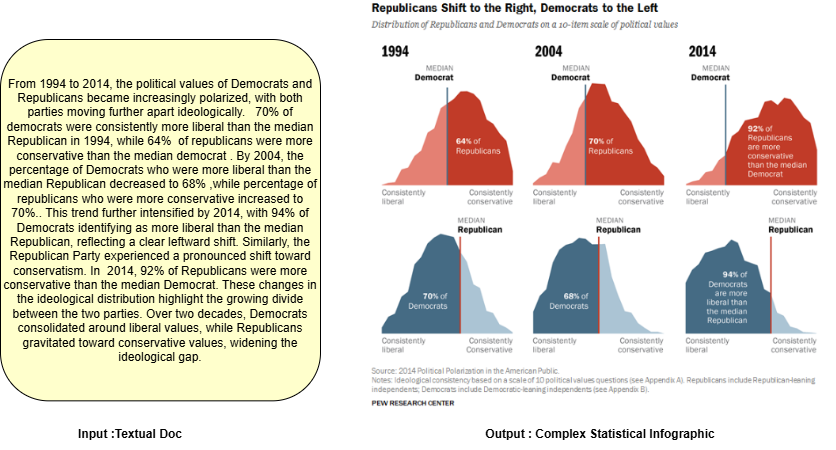}

   \caption{Sample instance of \textit{\textbf{text to complext statistical infographic}} task. The input is a textual document as shown in the left and the output is the corresponding statistical infographic in right.}
     \label{intro_figure}

\end{figure*}

\textbf{Research Gap:} Previous research has improved LLMs' ability to understand insights from infographics \cite{liu2022deplot, liu2022matcha, masry2023unichart}. Other studies have focused on creating simple statistical infographics from text and data \cite{dibia2023lida, tian2024chartgpt}, showing how AI can help with data visualization. However, these works mainly focus on basic charts like bar or line graphs based on structured instructions. Also, they take data in the form of CSV  and tabular formats.  However, if the user provides just the information in the form of textual documents, then generating the statistical infographic that consists of many sub charts requires a higher level of alignment and reasoning capabilities of LLMs. \par

\textbf{Motivation:} In this work, we aim to address this research gap by curating a benchmark dataset named \textbf{\textit{Infodat}}. Each datapoint of \textbf{\textit{Infodat}} consists of the textual document as input. The output is the metadata, consisting of all the information that will guide the final code generation for the infographic. Metadata here refers to structured information describing the characteristics of infographics, including chart types (e.g., line, bar), axis labels, statistical data points, text annotations, layout details (e.g., position, alignment, dimensions), and visual styles (e.g., fonts, background color). It organizes and contextualizes subcharts within an infographic, ensuring clarity and coherence in representation. Our primary motivation for generating intermediate metadata is inspired by works such as \cite{yan2023coco,lee2024instruction} demonstrating that providing explicit instructions to LLMs significantly improves output quality based on the use case. As there is no existing dataset for this task, we use a semi-automated approach for curating this dataset.\footnote{Examples of metadata are shown in Appendix A.6.} 

We propose a novel framework, \textbf{\textit{Infogen}}, with two main stages: the metadata generation module and the code generation module. The metadata generation module uses fine-tuned LLMs and a ranker to select the best output from these models. The code generation module includes LLM agents \cite{yang2024matplotagent}, namely the coder and feedback modules. The coder generates code from the metadata, while the feedback module refines the code based on the input document to improve the visual quality of the infographic. The goals of \textbf{\textit{Infogen}} are to accurately determine the number and types of sub-charts and their content from the document and to arrange them in a visually appealing and well-aligned format. A sample instance of our task is illustrated in Figure \ref{intro_figure}.\par

\textbf{Contributions}: We make four main contributions in this paper:\par

1) We introduce the \textbf{task} of text to complex statistical infographic generation.\par

2) We present a synthetic data curation approach to obtain text-to-infographic parallel data along with intermediate metadata. We refer to this resulting dataset as \textbf{\textit{Infodat}}.\par 

3) We propose \textbf{\textit{Infogen}}, a novel framework that performs content planning using LLMs to first generate intermediate metadata and then generate the final infographic and achieve state-of-the-art results for this task. \par

4) We propose a novel \textbf{evaluation setup} using various automatic metrics to evaluate the performance of different models for this task.\par

5) We conduct extensive empirical and qualitative analyses to measure our model's effectiveness for generating statistical infographics from text.\par

\section{Related Works}
\textbf{Statistical Infographics Understanding:} Research on this topic is broadly divided into two main approaches. The first involves using a single model to directly interpret charts and answer questions in natural language, as seen in works like \cite{liu2022matcha,masry2023unichart}, which streamlines the process by handling both visual and textual data simultaneously. The second approach converts charts into structured data for analysis, allowing for more accurate interpretations, as demonstrated by works like \cite{liu2022deplot,xia2023structchart}. Our approach is inspired by the second method, where we convert text data into structured metadata for final code generation.

\textbf{Works on Infographics Generation:} 
Recent work on statistical infographic generation has used LLMs to make the process easier and more automated. \citet{han2023chartllama} introduced ChartLlama, a multimodal LLM that understands and creates charts, performing well on chart comprehension tasks. ChartGPT \cite{tian2024chartgpt} turns natural language and data tables into simple charts. LIDA \cite{dibia2023lida} takes a step-by-step approach to create flexible visualizations and infographics with an easy-to-use interface for beginners. DataNarrative \cite{islam2024datanarrative} proposes a two-agent LLM framework that generates coherent data stories by integrating textual narratives with dynamically created visualizations. Socrates \cite{wu2023socrates}  introduces an adaptive system that creates data stories by eliciting and incorporating user feedback to iteratively improve narrative visualizations. DataShot \cite{wang2019datashot} presents a method for automatically generating fact sheets from tabular data, combining charts, icons, and key metrics into infographic-style summaries. Calliope \cite{shi2020calliope} develops a system that transforms spreadsheets into visual data stories composed of annotated charts and narrative text, enabling user-guided infographic editing.

\section{Development of \textit{Infodat} Dataset }
\label{sec:dataset}
To the best of our knowledge, no parallel dataset exists for generating complex infographics from text. To address this, we curate \textbf{\textit{InfoDat}} using a semi-automated approach to ensure high-quality synthetic data, leveraging the publicly available Pew dataset \cite{kantharaj2022chart}, which comprises statistical charts scraped from Pew Research articles.\footnote{Pew Research (www.pewresearch.org) publishes data-driven articles on social issues, public opinion, and demographic trends, featuring high-quality charts with professional descriptions.} The curation involves three steps: \textbf{Selection of Complex Statistical Infographics}, \textbf{Synthesizing Input Text Document}, and \textbf{Synthesizing Intermediate Metadata}.
\par
\textbf{Selection of Complex Infographics:} As mentioned in the introduction, this work focuses only on complex statistical infographics. Therefore, we use GPT-4o \cite{GPT4(V)} with few-shot prompting to filter out the simple charts from the Pew dataset. This results in 3,463 complex infographic images. The full prompt is provided in Appendix-\ref{sec:prmopts}\par

\textbf{Input Text Document Synthesis:} As there are no parallel text samples for the infographics, we synthetically curate them for the selected images. 
For this, we instruct GPT-4o \cite{GPT4(V)} using chain-of-thought prompting, along with a few examples, for the filtered statistical infographics. We ensure that no source image information or metadata details, such as the number of charts, are included in the final generated text. This is done to prevent any data leakage about the output metadata in the synthetic input text.
The full prompt is provided in Appendix-\ref{sec:prmopts}.\par

\begin{figure*}[hbt!]
    \centering
    \begin{subfigure}[b]{0.48\textwidth}
        \centering
        \includegraphics[height=4cm, width=7cm]{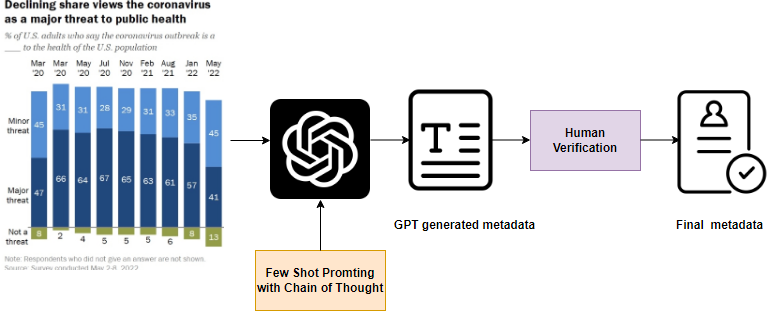}
        \caption{MetaData Generation Pipeline}
        \label{fig:metadata_pipeline}
    \end{subfigure}
    \hfill
    \begin{subfigure}[b]{0.48\textwidth}
        \centering
        \includegraphics[height=4cm, width=7cm]{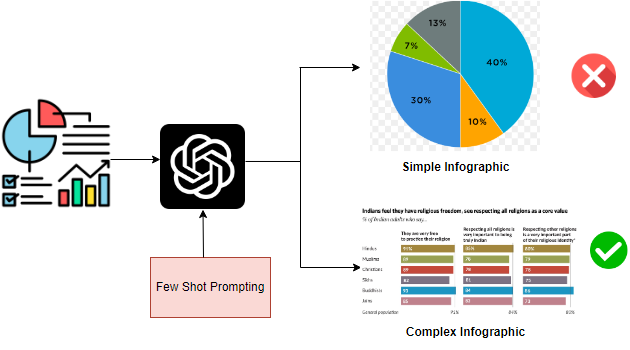}
        \caption{Infographic Selection Pipeline}
        \label{fig:infographic_pipeline}
    \end{subfigure}
    \caption{\footnotesize{The various stages of generating the \textit{\textbf{Infodat}} dataset are outlined as follows. We generated the metadata from the selected infographics using the pipeline illustrated in (a). Additionally, we identified the complex infographics by following the pipeline shown in (b).}}
    \label{fig:data_gen_pipeline}
\end{figure*}

\textbf{MetaData Synthesis:} We hypothesize that generating infographics based on intermediate, well-structured metadata is more effective than directly generating them from text. To create this metadata, we use a semi-automated approach. Metadata for filtered infographics is initially generated using few-shot prompting with Chain of Thought (COT) on GPT-4. To ensure accuracy and reliability, human reviewers refine and correct any errors based on the infographic images. Figure \ref{fig:data_gen_pipeline} illustrates the stages of generating the \textit{\textbf{Infodat}} dataset.
Table \ref{DStat} shows some statistics.\footnote{The human verification steps are included in Appendix FAQ Question-1.}
The prompts used for the construction of \textit{\textbf{Infogen}} dataset are given in Appendix -\ref{sec:prmopts}.


\begin{table}[hbt!]
    \centering
    \scalebox{0.65}{
    \begin{tabular}{lr}
    \toprule
       \textbf{Statistical Parameters} & \textbf{Value} \\
       \midrule
         \# of data points & 3,463 \\
         Avg. \# of words in metadata & 341.015 \\
         Median \# of words in metadata & 300 \\ 
         Avg. \# of words in input text & 185.46 \\
         Median \# of words in input text & 162 \\
         Avg. \# of sentences in metadata & 14.54 \\
         Avg. \# of sentences in input text & 15.07 \\
         Avg. \# of sub-charts in each metadata & 2.15 \\
         Median \# of sub-charts in each metadata & 2 \\
         Maximum \# of sub-charts in each metadata & 21 \\
         Minimum \# of sub-charts in each metadata & 1 \\
         \bottomrule
    \end{tabular}}
    \caption{\footnotesize{Statistical analysis of \textit{\textbf{Infodat}}.}}
    \label{DStat}
\end{table}

\section{Proposed Methodology}
This section outlines the problem and provides the foundation for understanding the proposed framework in detail.
\subsection{Problem Statement}
In our approach, we tackle the task of text-to-complex statistical infographic generation using \textbf{\textit{Infogen}}. The framework takes a document \( T \) containing statistical information in text form and processes it in two steps. First, \( f(T) \) converts the text into structured metadata \( M \), capturing elements like sub-charts, titles, and statistical details. Then, \( g(M) \) translates the metadata into executable Python code \( C \), ensuring proper alignment and aesthetics. The process is summarized as:

\[
C = g(f(T))
\]

\subsection{Text to MetaData Generation:} 
This can be broadly divided into three stages:\par

\textbf{ a) Fine-tuning of Large Language Models:} For the text-to-metadata generation task, we utilize large-scale models such as Qwen-2 Large \cite{yang2024qwen2}, LLAMA 3 \cite{dubey2024llama}, and Phi-3 Medium \cite{abdin2024phi}, leveraging efficient fine-tuning techniques like Quantized LORA \cite{dettmers2023qlora}. Although smaller models are tested, the larger variants consistently deliver superior accuracy and performance. The training objective is to minimize the cross-entropy loss:

\begin{scriptsize}
\textbf{
\begin{equation}
\mathcal{L} = - \sum_{i=1}^{N} y_i \log(\hat{y}_i)
\end{equation}
}
\end{scriptsize}

where \( y_i \) denotes the true metadata labels, and \( \hat{y}_i \) represents the model's predictions.

\par

\textbf{b)Alignment using  DPO:} To improve text-to-metadata generation, we align outputs from large-scale models—Qwen-2 Large (72B) \cite{yang2024qwen2}, LLAMA 3 (70B) \cite{dubey2024llama}, and Phi-3 Medium \cite{abdin2024phi}—using a method similar to Direct Preference Optimization (DPO) \cite{rafailov2024direct}. Since no preference dataset exists for this task, we create a synthetic one using GPT-3.5 Turbo, inspired by \cite{ghosh2024healthalignsumm}. For each data point, we generate two metadata outputs from our fine-tuned models (using varying temperatures) and have GPT-3.5 Turbo rank them as preferred (\( y_w \)) or less preferred (\( y_l \)). This dataset is then used to fine-tune the models with DPO.

The DPO loss, \( L_{\text{DPO}}(\pi_\theta; \pi_{\text{GPT}}) \), is calculated over the synthetic dataset \( D_{\text{synthetic}} \):

\begin{scriptsize}
\textbf{
\begin{multline}
L_{\text{DPO}}(\pi_\theta; \pi_{\text{GPT}}) = \\
- \mathbb{E}_{(x, y_w, y_l) \sim D_{\text{synthetic}}} \left[ \log \sigma \left( \beta \log \frac{\pi_\theta(y_w \mid x)}{\pi_{\text{GPT}}(y_w \mid x)} \right. \right. \\
\left. \left. - \beta \log \frac{\pi_\theta(y_l \mid x)}{\pi_{\text{GPT}}(y_l \mid x)} \right) \right]
\end{multline}
}
\end{scriptsize}

Here, \( \pi_\theta(y_w \mid x) \) and \( \pi_\theta(y_l \mid x) \) are the probabilities assigned by our model to the preferred and less preferred outputs, and \( \pi_{\text{GPT}}(y_w \mid x) \), \( \pi_{\text{GPT}}(y_l \mid x) \) are the probabilities assigned by GPT-3.5 Turbo. The scaling factor \( \beta \) and the sigmoid function \( \sigma \) enhance the preference contrast. This method aligns the model's outputs, improving metadata generation quality. Details on the prompt for the synthetic dataset are provided in Appendix \ref{sec:prmopts}.\par

\textbf{c) Ranker LLM:} We introduce a ranker to improve metadata generation accuracy by addressing situations where the top-performing model may sometimes produce incorrect hallucinated outputs \cite{sahoo2024unveiling}. The ranker evaluates outputs from multiple models to identify and select the most accurate result, ensuring better overall performance. The ranker, fine-tuned using LLAMA 3 (70B), evaluates outputs from three DPO-fine-tuned LLMs. It takes the context and three options as input and ranks them to identify the most accurate and suitable output. The ranker is trained on a dataset created from the outputs of DPO-fine-tuned models. We apply heuristics, such as verifying the number of sub-charts and ensuring the correct sub-chart types, to evaluate these outputs. This dataset enables the ranker to reliably choose the best output from multiple LLMs.
\footnote{The prompt used for ranker LLM has been stated in Appendix \ref{sec:prmopts}.} \par

\begin{figure*}[hbt!]
    
    \centering
    
    \includegraphics[width=14cm, height=9cm]{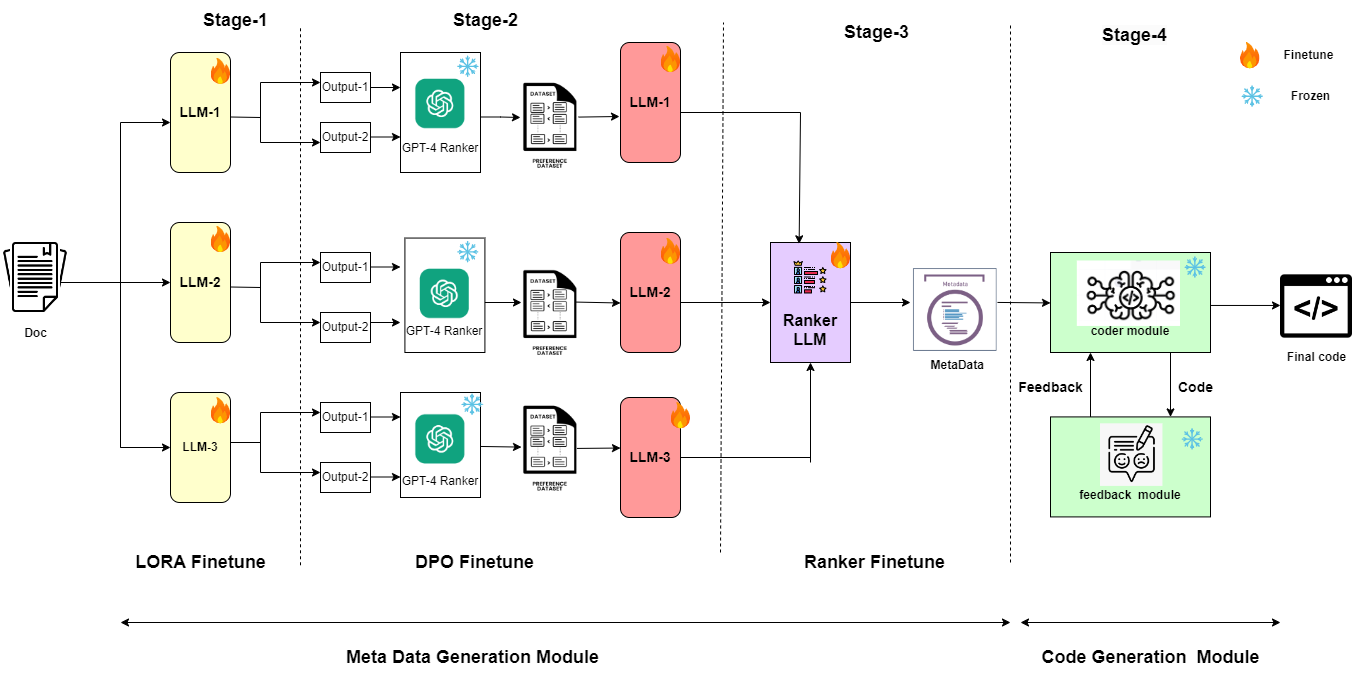}
    \setlength{\abovecaptionskip}{10pt}
  \setlength{\belowcaptionskip}{-2pt}
    
    \caption{\footnotesize{The architecture of our proposed framework, \textit{Infogen}, consists of two main stages: the \textit{MetaData Generation Module} and the \textit{Code Generation Module}. The \textit{MetaData Generation Module} is broken down into three steps: first, fine-tuning LLMs; second, post-training with synthetic data generated using DPO; and finally, using a fine-tuned LLM as a ranker to produce the final metadata. In the \textit{Code Generation Module}, there are two LLM agents: the coder agent, which generates code from the metadata, and the feedback module, which reviews and provides feedback on the generated code based on the metadata.}}
     \label{model_architecture}

\end{figure*}

\subsection{MetaData to Final Infographic Generation:}
Our approach for generating final code from metadata involves two LLM modules: the \textbf{Coder Module} and the \textbf{Feedback Module}. Since the coder module may not produce fully functional code initially, the feedback module reviews the metadata and generated code, providing suggestions for improvement. These modules work iteratively to refine the code, as detailed below.\par 
\textbf{Coder Module:}
The first stage converts the metadata into executable code using a large language model, such as GPT-4o \cite{GPT4(V)}. This metadata, containing details like subchart types, axes, statistical data, and layout, is processed by the coder module. Through in-context learning, the model generates code that includes: (1) Subchart Setup – defining subchart types and properties; (2) Data Integration – mapping statistical data to subcharts; and (3) Layout Structuring – arranging subcharts cohesively. This ensures the code reflects the metadata's structure and data. Advanced Python libraries like Plotly \cite{plotly_line_charts}, and Plotnine \cite{plotnine_website} are used for code generation.\par

\textbf{Feedback Module:}
After generating the initial code, a refinement stage improves its accuracy and alignment with the input text. The feedback module reviews the code for issues like incorrect data mappings, sub-chart properties, or layout inconsistencies. It provides suggestions that the coder module uses to refine the code iteratively, up to five cycles, to ensure high-quality output. This two-stage process—initial generation and iterative refinement—ensures the final code is accurate and aligns with the metadata and input context. The prompts are detailed in Appendix \ref{sec:prmopts}, and the complete \textit{\textbf{Infogen}} framework is shown in Figure-\ref{model_architecture}.

\begin{table*}[ht]
    \centering
    \scalebox{0.5}{ 
    \renewcommand{\arraystretch}{1.2}
    \begin{tabular}{|l|c|c|c|c|c|c|c|c|}
    \hline
    Model Configuration & Subchart Accuracy & RSE & Title Rouge-L & Summary Rouge-L & Subchart Type Accuracy & Subchart Summary Rouge-L & Statistical Accuracy \\ \hline
    GTP4o\_10\_shot & 52.28 & 2.43 & 0.34 & 0.29 & 68.96 & 0.40 & 87.34 \\ 
    GTP4o\_20\_shot & 56.73 & 2.06 & 0.36 & 0.31 & 72.10 & 0.42 & 87.77 \\ 
    GTP4o\_BM25\_clustering\_10shot & 55.76 & 2.10 & 0.37 & 0.33 & 78.03 & 0.44 & 88.44 \\ 
    \textbf{GTP4o\_BM25\_clustering\_20shot} & \textbf{57.69} & \textbf{2.20} & \textbf{0.39} & \textbf{0.35} & \textbf{79.07} & \textbf{0.43} & \textbf{88.62} \\ \hline
    reft\_llama & 52.5 & 2.25 & 0.38 & 0.36 & 66.82 & 0.39 & 74.28 \\ 
    reft\_phi3 & 50.96 & 2.12 & 0.36 & 0.34 & 68.22 & 0.37 & 68.94 \\ 
    \textbf{reft\_qwen2} & \textbf{49.80} & \textbf{2.25} & \textbf{0.46} & \textbf{0.42} & \textbf{70.57} & \textbf{0.44} & \textbf{79.19} \\ \hline
    llama3\_qlora\_small & 53.96 & 2.53 & 0.55 & 0.39 & 81.92 & 0.45 & 85.72 \\ 
    llama3\_full\_small & 59.6 & 2.21 & 0.49 & 0.41 & 79.61 & 0.42 & 79.15 \\ 
    llama3\_qlora\_large & 66.11 & 2.13 & 0.55 & 0.48 & 82.58 & 0.51 & 89.37 \\ 
    \textbf{llama3\_qlora\_large\_dpo} & \textbf{68.65} & \textbf{2.05} & \textbf{0.55} & \textbf{0.48} & \textbf{82.98} & \textbf{0.51} & \textbf{88.27} \\ \hline
    phi3\_qlora\_small & 37.69 & 2.08 & 0.51 & 0.46 & 80.96 & 0.50 & 85.38 \\ 
    phi3\_qlora\_large & 63.46 & 1.96 & 0.53 & 0.47 & 82.78 & 0.51 & 88.65 \\ 
    \textbf{phi3\_qlora\_large\_dpo} & \textbf{72.11} & \textbf{1.96} & \textbf{0.56} & \textbf{0.40} & \textbf{83.03} & \textbf{0.51} & \textbf{89.44} \\ \hline
    qwen\_qlora\_small & 45.96 & 2.70 & 0.45 & 0.42 & 6.60 & 0.03 & 6.61 \\ 
    qwen\_qlora72B\_large & 68.26 & 2.04 & 0.50 & 0.43 & 81.92 & 0.49 & 87.47 \\ 
    \textbf{qwen\_qlora\_large\_dpo} & \textbf{71.73} & \textbf{1.94} & \textbf{0.52} & \textbf{0.49} & \textbf{80.09} & \textbf{0.49} & \textbf{88.11} \\ \hline
    Incontext\_learning\_with\_LLMs\_merge & 65.57 & 2.24 & 0.51 & 0.39 & 83.46 & 0.49 & 88.79 \\
    \textit{\textbf{Infogen (small)}} & \textbf{59.2} & \textbf{2.25} & \textbf{0.55} & \textbf{0.40} & \textbf{84.5} & \textbf{0.47} & \textbf{87.2}
    \\
     \textit{\textbf{Infogen (large)}} & \textbf{74.69} & \textbf{1.80} & \textbf{0.56} & \textbf{0.49} & \textbf{84.23} & \textbf{0.52} & \textbf{89.56} \\ \hline
    \end{tabular}
    }
    \caption{\footnotesize{The performance comparison of different model configurations for the text-to-metadata generation task on the \textbf{\textit{Infodat}} dataset shows that \textbf{\textit{Infogen (large)}} outperforms all baselines. Among the prompting models, GPT-40 with BM25 clustering performs the best, while Phi3 (large) with DPO leads among the fine-tuned LLMs. The best-performing model in each category is highlighted in bold.}}
    \label{tab:model_metrics}
\end{table*}

\section{Experimental Results and Analysis}

This section details the experimental setup and comprehensive evaluation of the proposed model, \textit{\textbf{Infogen}},'s performance using automatic, human, and qualitative evaluation.

\subsection{Experimental Setup}

Our experiments were conducted on an A100 80GB GPU machine, with each model requiring an average runtime of around 5 to 6 hours. We utilized the PyTorch \cite{paszke2019pytorch}, Hugging Face \cite{wolf2019huggingface}, and unsloath \cite{unsloth_website} libraries to implement both the baselines and our proposed architecture.  The dataset was divided into training, validation, and test sets in an 80:5:15 ratio. Key hyperparameters for the metadata generation module included a per-device training batch size of 2, gradient accumulation steps of 4, warmup steps of 5, a maximum of 524 steps, a learning rate of 2e-4, an AdamW 8-bit optimizer, weight decay of 0.01, 
and a linear scheduler.  We used LLAMA-3 (80GB) calls with max\_tokens of 1000 and temperature of  0.5 for the code and feedback module. \par

\subsection{Data Preprocessing}
 We utilize the \textit{\textbf{Infodat}} dataset, which was curated through a semi-automated process involving the selection of complex statistical infographics, the synthesis of corresponding input text, and the generation of structured metadata. The dataset was split into training, validation, and test sets with an 80:5:15 ratio. As part of preprocessing, the input text for each infographic was carefully crafted to avoid data leakage, ensuring that no source image metadata was included in the text to maintain the integrity of the task.\par 

\subsection{Baselines Setup}

For the text-to-metadata task, we conducted a comprehensive evaluation with various baselines, including few-shot prompting using state-of-the-art models like GPT-4o. To enhance diversity in few-shot prompts, we employed BM25 embeddings for clustering and selected examples from cluster centroids, referred to as \textbf{GPT4o\_BM25\_clustering} in Table \ref{DStat}. Additionally, we explored fine-tuning smaller parameter models with REFT \cite{wu2024reft} and evaluated state-of-the-art LLMs like Qwen2, LLAMA3, and Phi3 (both smaller and larger versions). Finetuning was also performed using alignment techniques like DPO with synthetic preference datasets (generated with GPT-4o as a judge). Lastly, we tested a mixture of LLM agents using in-context learning, denoted as \textbf{Incontext\_Learning\_withLLMsmerge}. More details of the baselines are provided in Appendix \ref{sec:baselines}.

\par
\subsection{Evaluation Metrics}
As there were no available metrics for this task, we introduce the following measures for evaluating the text-to-metadata generation task:
\textbf{(a) Sub-chart Accuracy:}  Percentage of the number of sub-charts in the generated metadata compared to the ground truth, RSE is the root square mean error for the number of sub-charts in the ground truth with respect to the generated metadata.
\textbf{(b) Sub-chart Type Accuracy:} Percentage of correct sub-chart type classifications.
\textbf{(c) Statistical Accuracy:} Percentage of statistical numbers correctly extracted from the text in the metadata.
\textbf{(d) Textual Information Metrics:} Rouge-L scores for the generated titles, summaries, and sub-chart-level summaries in the metadata \cite{lin2004rouge,ghosh2024sights}.
Further details for each of these metrics are provided in Appendix section \ref{sec:metrics}.
For the final infographics, we conduct a human evaluation focusing on readability, visual appeal, and data accuracy  based on the metadata.

\begin{table}[h]
\centering
\scalebox{0.4}{
\renewcommand{\arraystretch}{2.5}
\begin{tabular}{|c|c|c|c|}
\hline
\textbf{Model}     & \textbf{Readability Score} & \textbf{Visual Appeal Score} & \textbf{Data Accuracy and  Alignment Score}  \\ \hline
GPT-4o (20 shot)   &  3.4       &     2.8     &   2.4                       \\
Phi3(DPO)  &  3.7     &    3.2      &   3.4   \\
\textit{\textbf{Infogen}} &   \textbf{4.1}      &    \textbf{3.8}     &    \textbf{4.1}  \\ \hline
\end{tabular}
}
\caption{\footnotesize{Human evaluation of \textbf{\textit{Infogen}} with respect to different baselines.}}
\label{tab:humane}
\end{table}

\begin{figure*}[hbt!]
    
    \centering
    
    \includegraphics[width=14cm, height=7cm]{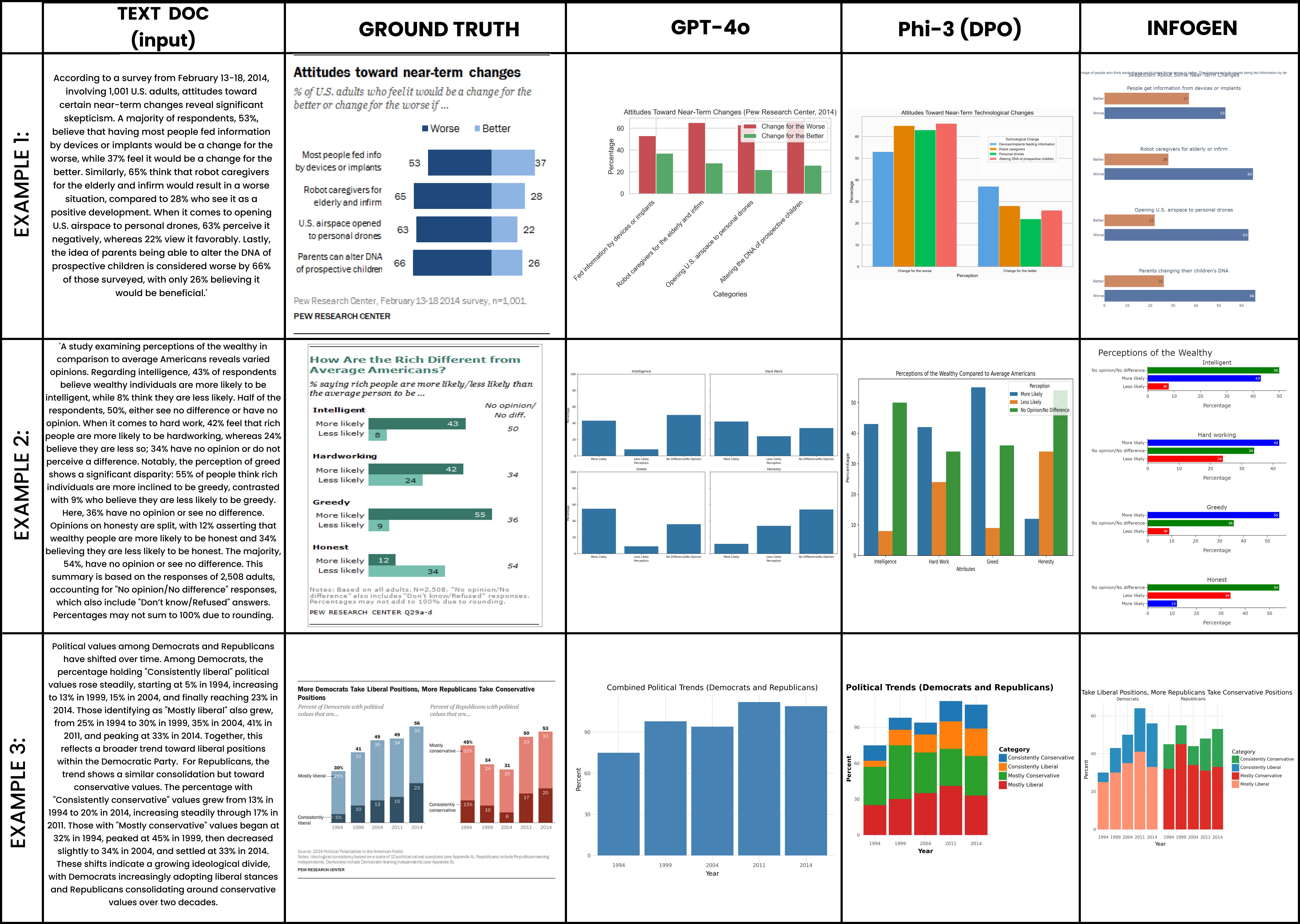}
    \setlength{\abovecaptionskip}{10pt}
  \setlength{\belowcaptionskip}{-2pt}
    
    \caption{\footnotesize{The qualitative analysis of the generated infographics from GPT-4o, Phi3 (DPO), and \textit{Infogen} reveals a clear advantage of \textit{Infogen}. Notably, \textit{Infogen} accurately identifies the correct number of sub-charts from the text and captures the alignment of the sub-charts perfectly, outperforming the other two models in these aspects.}}.
    \label{qualitative_analysis}

\end{figure*}

\section{Results and Findings}
Table-\ref{tab:model_metrics} presents a comprehensive analysis of the task of text-to-metadata generation, comparing various model configurations with our proposed architecture, \textit{\textbf{Infogen}}. Building on our hypothesis that improved metadata leads to the creation of better infographics, the extensive experiments address the following research questions in detail.
\par
\subsection{Research Questions}
\textbf{R1)  How does \textit{\textbf{Infogen}} perform in comparison to existing prompting and fine-tuning strategies using LLMs? }  
\textit{\textbf{Infogen}} achieves state-of-the-art performance in generating complex statistical infographics from text, with the highest scores in \textbf{Subchart Accuracy} (74.69\%) and \textbf{Statistical Accuracy} (89.56\%), outperforming other models. These results confirm its effectiveness for text-to-infographic tasks. Moreover, fine-tuning Large Language Models (LLMs) consistently outperforms prompting-based methods, highlighting the critical role of fine-tuning in achieving superior performance.\par 
\textbf{R2) Does alignment algorithms like Direct Preference Optimization (DPO)  help in improving the accuracy of this generation?}  
Fine-tuning Large Language Models (LLMs) using techniques like Direct Preference Optimization (DPO) leads to much better results. For example, Phi-3 (DPO) performs far better than its version without DPO. Similar improvements are seen in other models like LLAMA3 and Qwen-2, showing that DPO is very effective at improving the accuracy of generated infographics. This shows that even preferences generated by models like GPT-4o are good enough to significantly improve the performance of LLMs for tasks like text-to-metadata generation, making alignment methods like Direct Preference Optimization very powerful.\par
\textbf{R3)  What is the influence of the feedback module in the final generated infographic?}
The feedback module reviews Python code from the coder module to ensure it aligns with the metadata. It verifies the number and type of subcharts, checks axes, statistics, and subchart positioning, and evaluates alignment, layout, fonts, and dimensions. It also ensures the title and summary are included, with no extra axes or errors, ensuring accurate, well-formatted visualizations. Figure-\ref{fig:importance_feedback} illustrates how it helps the coder module correct overlapping text errors.

\textbf{R4)  How robust is \textit{\textbf{Infogen}}? Does having multiple fine-tuned LLMs help in the metadata generation?}   To show the robustness of \textit{\textbf{Infogen}}, we use the same framework with smaller versions of Phi-3 mini, LLAMA 3 (8B), and Qwen-2 (7B) versions. Even with smaller LLMs, \textbf{\textit{Infogen}} delivers better results than individual small models. For example, \textbf{\textit{Infogen (small)}} achieves better performance in metrics like \textbf{sub-chart accuracy} ($59.2\%$) compared to individual models such as \textbf{reft\_llama} and \textbf{phi3\_qlora\_small}. This demonstrates the robustness of \textbf{\textit{Infogen}}, suggesting that while a system with multiple LLM agents can further enhance performance, even a simplified version of \textbf{\textit{Infogen}} with smaller models still outperforms standalone models. 
In scenarios where inference time is more important than peak performance, a smaller version, \textit{\textbf{Infogen (small)}}, can be used. 

\par
\vspace{-0.3cm}
\subsection{Ablation Analysis}
\textbf{Performance of Incontext Learning in comparison to  Finetuned Models :} We employed GPT-4o for in-context learning using the few-shot prompting technique. As baselines, we used randomly selected 10-shot and 20-shot examples. While a slight improvement was observed as the number of examples increased, performance improved further when in-context examples were selected using BM25 clustering. However, we saw a significant boost in performance when fine-tuned LLMs were used, particularly with larger models, a trend that consistently held across all LLMs evaluated.

\textbf{Small vs Large LLMs:} We perform extensive experiments with LLMs of different sizes, including LLAMA 3.1, Phi-3, and Qwen-2, and also tested their REFT fine-tuned versions. The results show that fine-tuning greatly improves the performance of these models, with the larger versions showing the most significant gains. This suggests that, for tasks where performance is a priority, using larger LLMs is the best approach, as they consistently deliver better results after fine-tuning.

\textbf{Selection of Best Ranker:} To determine the optimal ranker configuration, we conducted experiments using two approaches: few-shot prompting, referred to as \textbf{Incontext\_learning\_with\_LLMs\_merge}, and fine-tuned LLMs referred as \textit{\textbf{Infogen}}. Our findings highlight that the strength of our framework heavily depends on the performance of the ranker. Through these experiments, we observed that fine-tuned LLMs consistently outperformed the few-shot prompting approach for the ranker module. So for our proposed framework, \textit{\textbf{Infogen}}, we have used finetuned LLMs.

\subsection{Human Evaluation}

A team of annotators who have knowledge of statistical infographics conducted a human evaluation on 35\% of test samples randomly selected for this purpose. The evaluation metrics used:\textbf{Readability Score}, were provided to assess the clarity, organization, and flow of the generated infographic; \textbf{Visual Score}, which evaluated the layout, design, colors and subchart identification with respect to ground truth; and \textbf{Data Accuracy and  Subchart Alignment Score}, measuring the correctness of the data and the alignment of the sub-charts with respect to each other in the statistical infographic in 
effectively conveying key insights. Each of these metrics was rated on a scale from 1 to 5, with 1 representing the lowest quality and 5 indicating the highest quality of the generated infographics. We evaluated the generated infographics results of GPT-4o (20 shot), Phi3-large (DPO), and our proposed  \textit{\textbf{Infogen}}. The human evaluation results are shown in Table-\ref{tab:humane}. We have elaborated  each of these metrics in depth in the Appendix section \ref{sec:metrics}.

\vspace{1cm}

\vspace{-0.9cm}

\subsection{Qualitative Analysis of Generated Infographics}
Figure-\ref{qualitative_analysis} presents a comparative analysis of infographics generated by \textbf{GPT-4o}, \textbf{Phi-3 (DPO)}, and \textit{\textbf{Infogen}}, highlighting two key aspects: (1) \textit{\textbf{Infogen}} accurately captures statistical values by correctly mapping them to their respective subcontexts, ensuring contextual consistency; and (2) \textit{\textbf{Infogen}} effectively maintains alignment across all subcharts, achieving a cohesive and well-structured layout. In comparison, outputs from \textbf{GPT-4o} and \textbf{Phi-3 (DPO)} are cluttered and less readable, reducing their effectiveness in presenting insights.

\section{Conclusion}
This paper introduces the task of generating complex statistical infographics from text-heavy documents and presents the benchmark dataset \textit{\textbf{Infodat}} along with the framework \textit{\textbf{Infogen}}. \textit{\textbf{Infogen}} uses a two-stage process of metadata and code generation to convert unstructured text into well-aligned, multi-subchart infographics, achieving state-of-the-art performance. By leveraging fine-tuned large language models and iterative feedback, it ensures high accuracy and visual appeal. Future directions include expanding to healthcare and finance, enabling template customization, and improving efficiency with faster inference techniques.

\section{Limitations}  
There are some noticeable limitations of our work. They are enumerated in the points below:\par

1)Though \textit{\textbf{Infogen}} delivers state-of-the-art performance, there is room for improvement in accurately determining and aligning sub-charts. Misalignments can cause infographics to miss key data structure details, reducing their effectiveness. Addressing this would enhance the clarity and reliability of the visuals.\par

2) Our dataset, \textit{\textbf{Infodat}}, is relatively limited in size, which might affect the generalizability of the model in diverse domains like healthcare. \par

3) The framework currently lacks support for customized template selection based on context, limiting its flexibility in tailoring infographics for specific user needs or varied contexts. Future improvements would focus on integrating customizable templates to enhance the user experience and adaptability across different use cases.\par

\section{Ethics Statement}
 
We utilized publicly available data for this study, specifically \textit{\textbf{Infodat}}, which was curated following a semi-automated approach for the generation of statistical infographics. No personally identifiable information (PII) was involved in this dataset, and all source information has been removed from both the input document and metadata, and all data points were derived from publicly available statistical datasets. Any sensitive information that could potentially be included in the textual documents used for metadata generation was redacted to prevent privacy violations. The dataset does not contain harmful or offensive content, and all experiments were conducted in accordance with ethical research standards. The use of this dataset is strictly for academic and research purposes, ensuring that no personal data or harmful material was processed during this work.

\section{Acknowledgement}
We thank the ACL ARR reviewers for their constructive comments and valuable feedback on the draft. Also we would like to thank all the annotators for verification of the dataset used in this project. Akash would like to thank Adobe Research for their support and for providing all the necessary resources during his internship in the summer of 2024, which greatly contributed to the success of this research project.

\bibliography{MMCQS}
\bibliographystyle{acl_natbib}

\appendix

\section{Appendix}
The Appendix section covers FAQs, risk analysis , statistical analysis of \textit{\textbf{Infodat}} ,
examples of metadata along with corresponding infographics and prompts used for this work.

\subsection{Frequently Asked Questions (FAQs)}

\textbf{1.  What are the human annotation guidelines for synthetic metadata verification ?}\par
Ans: We instruct the human annotators to follow the below guidelines to verify the accuracy of metadata generated from images containing multiple sub-charts :

\par \textbf{a. Count the Total Number of Subcharts:} 
Ensure the metadata correctly identifies the total number of subcharts in the image. Count the subcharts in the image and compare them with the metadata.(Each subchart should cover a different context of information.)

\par \textbf{b. Identify the Type of Each Subchart:} 
For each subchart, verify the type (e.g., bar chart, line chart, pie chart). Ensure that the type specified in the metadata matches the actual subchart in the image after matching the corresponding context.

\par \textbf{c. Describe the Axes:} 
Check the axes of each subchart. Verify that the metadata correctly describes the axes, including labels, units, and the type of data (e.g., percentage, time, categories). Make sure both X and Y axes are accurately represented.

\par \textbf{d. Extract Key Statistics:} 
Review the data points or statistics shown in each subchart. Compare the data with what’s provided in the metadata and ensure that it matches.

\par 

\textbf{e. Verify Subchart and Text Position:}
Ensure the metadata accurately describes the relative position of each subchart and its associated text (e.g., "top-left", "below the first subchart", "text above the subchart"). Verify that both the subchart and text are correctly placed as described.

\par
\textbf{2. What are the annotator demographics and what was the inter-annotator agreement?}\par  
Ans: The annotators comprised graduate interns with expertise in statistical infographics and prior experience with Pew/statistical data. For the final evaluation of the generated infographics, the inter-annotator agreement was measured using Cohen’s Kappa, yielding a score of 0.78.\par
\textbf{3. Why ChartGPT and Lida are not taken into baselines ?}\par
Ans: ChartGPT and LIDA operate in a setup where both text and accompanying data tables (e.g., CSV files) are provided as inputs. LIDA explicitly transforms a CSV file into an infographic, converting it into text internally, while ChartGPT uses a query and a CSV file to generate infographics step by step. In contrast, \textit{\textbf{InfoGen}} addresses a fundamentally different problem space by generating infographics directly from a text passage without relying on any external data tables or structured queries, focusing solely on unstructured text as input. \par

\textbf{4. Are the results statistically significant ?}\par
Ans: The results of Infogen both small and large are statistically significant The scores of them are the mean of three runs conducted.\par
\textbf{5. Is inference time  a bottleneck for \textit{\textbf{Infogen}} ?}\par
Ans:Yes, as \textbf{Infogen} is using both bunch of LLM agents in both metageneration and codegeneration phase . On avg \textbf{Infogen} takes 1.5 times more inference time . So, performance versus inference tradeoff is there.  We believe we can reduce the inference time by using recent techniques like Speculative decoding but the efficiency part of the framework is out of the scope of this work.\par

\textbf{6. How does the loop between the coder module and feedback module end?}\par
Ans: The loop ends when the judge module returns a "yes," indicating that the coder module has successfully generated the final code. If the judge module returns "no," the process repeats, with a maximum of 5 iterations.\par

\textbf{7.Can I see more examples of Dataset ?}\par

Ans: Please go through this anonymous GitHub account for the same: \url{https://anonymous.4open.science/r/Infogen_Samples-18C3/}.

\par
\textbf{8. What is the significance of this study compare to previous research works in this direction?}\par
Ans: This work represents the first attempt at generating complex statistical infographics that consist of multiple subcharts, focusing specifically on testing the capability of large language models (LLMs) in planning and reasoning purely from text documents. Unlike prior approaches, which rely on both text and data tables for infographic generation, our method challenges the LLMs to independently interpret and generate detailed visualizations using only the textual input.\par
\textbf{9. Will the same framework work for other domains like Healthcare and Finance?}\par
Ans: We are confident that the same framework can be applied to data from other domains. However, we believe that the training dataset should ideally include some domain-specific data before deploying it in new areas, as the metageneration phase relies on fine-tuned LLMs. Additionally, for certain domains, such as healthcare, the \textit{\textbf{Infogen}} framework should incorporate appropriate safety checks to ensure compliance and reliability.
\par

\textbf{10. Do you not feel the size of \textit{\textbf{Infodat}} dataset is rather small?}\par
Ans: In this work, we focus on incorporating infographics with multiple subcharts, deliberately excluding simpler ones to preserve the dataset's complexity and ensure meaningful evaluations. To facilitate broader usability, we provide detailed steps for extending the dataset, making it adaptable to various domains. Our data generation pipeline is designed to be robust, supporting seamless expansion. However, human validation of the metadata remains crucial to ensure it meets the expected quality standards. \par

\textbf{11. How the annotators and human verifiers compensated for their work ?}of metadata is must to check if its satisfy the expected quality .\par

Ans: They are compensated according to our agreement, which adheres to the Government's Minimum Wage guidelines.\par

\textbf{12 . Why were different models optimized on GPT-4.0’s preferences used instead of directly relying on GPT-4.0?}\par
Ans: We initially experimented with GPT-4.0 alone, employing configurations such as random 10- or 20-shot examples and BM25 clustering to select optimal examples. However, our results showed that larger white-box LLMs, such as LLAMA 3.1, Phi3, and Qwen-2, consistently outperformed GPT-4.0 in terms of accuracy. These models were further fine-tuned using DPO with GPT-4.0 preferences, significantly boosting their performance. Consequently, we integrated these models into the Infogen framework rather than relying solely on GPT-4.0. Table 1 provides a detailed comparison of our findings .\par

\textbf{13. Does the definition of infographics in the paper align with the past literature?}\par
Ans: In this study, we focus exclusively on the domain of statistical infographics, specifically targeting complex statistical infographics that consist of multiple subcharts, each with a distinct context.

\textbf{14. Is code generation evaluated independently?}\par

Ans: We acknowledge that code generation was not explicitly evaluated on its own. However, we extensively assessed the quality of the infographics generated from the code through human and qualitative evaluations. Our hypothesis, supported by previous works \cite{yan2023coco}, is that better code leads to higher-quality infographics, as demonstrated in Figure-\ref{qualitative_analysis}.

 \textbf{15. What aspects are covered by the metadata, and what are excluded?}

 Ans: The metadata captures structured details about infographics, including chart types (e.g., line charts), axis labels (e.g., months, percentages), statistical data points, text annotations, layout information (e.g., position, alignment, dimensions), and visual styles (e.g., fonts, background color). It organizes subcharts within the infographic, specifying their relationships and context, while ensuring clarity and coherence. The metadata does not explicitly address interactive elements, or advanced visual features like animations.\par

\subsection{Risk Analysis}

 Our approach, while effective, presents several risks. Misalignment between metadata and visuals can lead to infographics that misrepresent data or key insights. Additionally, the use of large language models may cause inconsistent outputs, especially with ambiguous input, affecting clarity and accuracy. Although feedback loops help refine results, human oversight is still crucial to catch potential errors, particularly in sensitive areas like healthcare or finance.

\subsection{Influence of Feedback to enhance the quality of generated infographics}

\begin{figure*}[hbt!]
    \centering
    \begin{subfigure}[b]{0.48\textwidth}
        \centering
        \includegraphics[height=4cm, width=7cm]{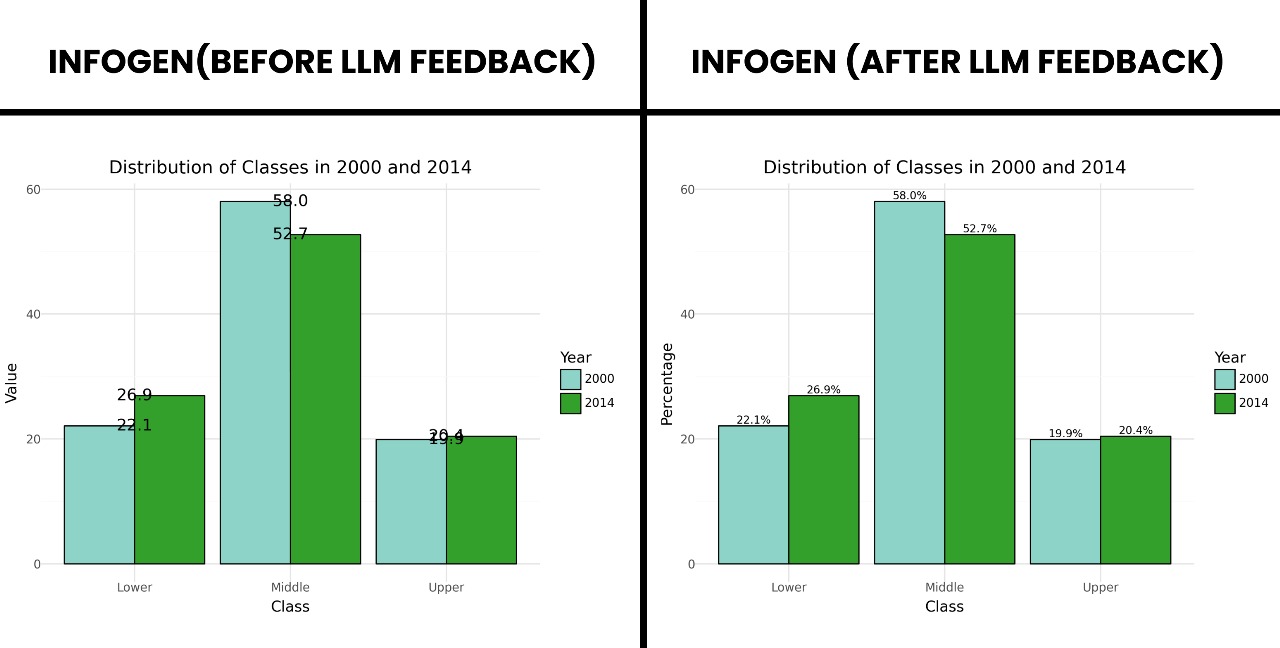}
        \caption{Example-1}
        \label{fig:metadata_pipeline}
    \end{subfigure}
    \hfill
    \begin{subfigure}[b]{0.48\textwidth}
        \centering
        \includegraphics[height=4cm, width=7cm]{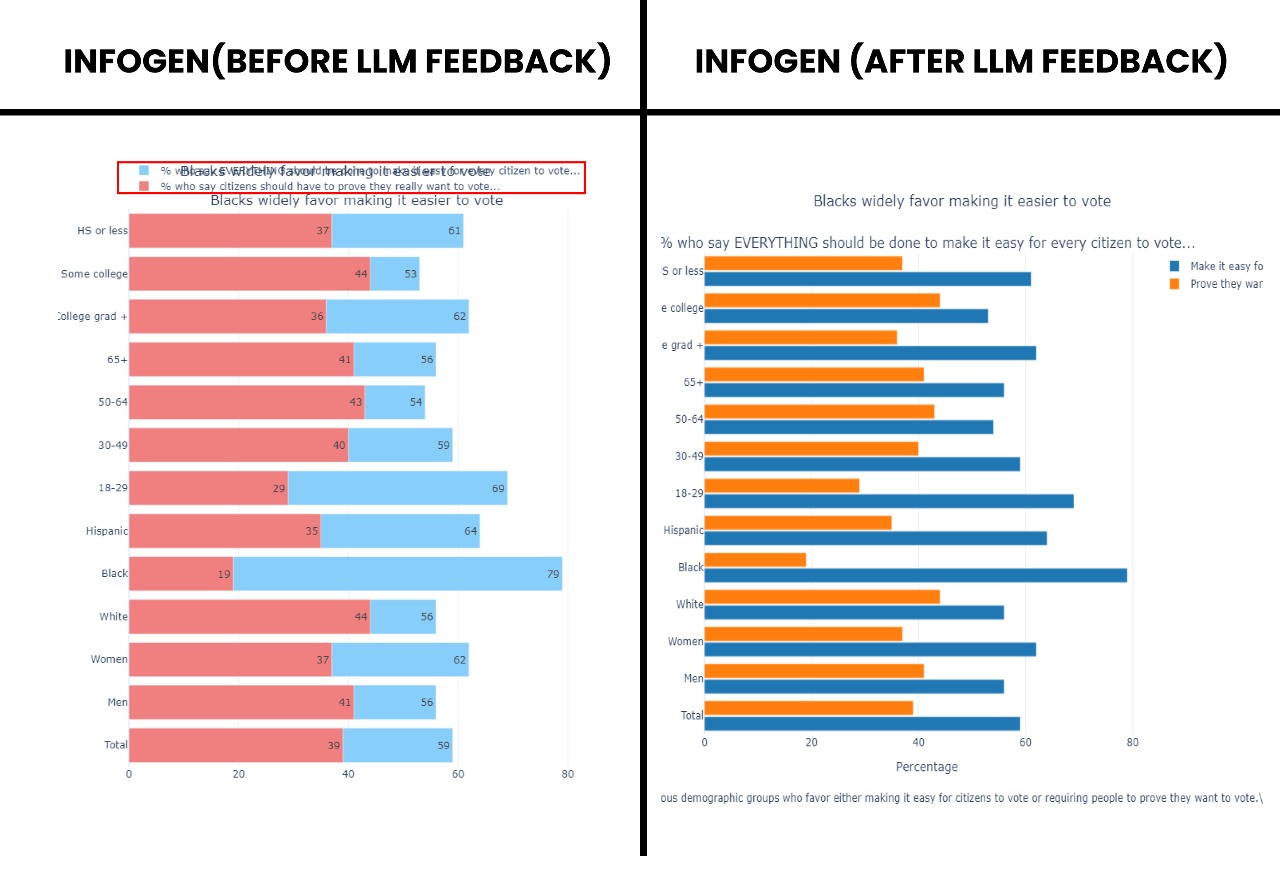}
        \caption{Example-2}
        \label{fig:infographic_pipeline}
    \end{subfigure}
    \caption{This figure demonstrates the impact of the feedback module on improving infographic generation. On the left, text overlap issues are evident in the initial output, but these are resolved after applying the feedback module. In Example 1, numerical values colliding with the bar are corrected, ensuring better clarity. Similarly, in Example 2, overlapping text above the horizontal bars is adjusted, resulting in a clean and well-aligned presentation.}
    \label{fig:importance_feedback}
\end{figure*}

The feedback LLM plays a pivotal role in the Infogen pipeline by ensuring the accuracy and quality of the generated code. After the initial code generation, the feedback module reviews issues such as incorrect data mappings, layout inconsistencies, or sub-chart misalignments. It provides iterative suggestions, enabling the coder module to refine the code over multiple cycles. This feedback process ensures the generated infographics adhere to metadata constraints, including subchart count, type, alignment, and visual appeal. By integrating this iterative feedback mechanism, the Infogen pipeline achieves high-quality outputs that align with the input text and metadata. Figure-\ref{fig:importance_feedback}
shows how feedback is helping to rectify the errors in the code generation.

\subsection{More details about Baselines}
\label{sec:baselines}

\begin{enumerate}

    \item \textbf{GPT-4o\_10\_shot and GPT-4o\_20\_shot}:
    \begin{itemize}
        \item These baselines involve using OpenAI's GPT-4o model with few-shot learning.
        \item In the 10-shot and 20-shot configurations, 10 and 20 examples respectively are provided to the model as in-context demonstrations during metadata generation.
        
    \end{itemize}
    
    \item \textbf{GPT-4o\_BM25\_clustering\_10shot , GPT-4o\_BM25\_clustering\_20shot}:
    \begin{itemize}
        \item These configurations enhance the few-shot prompting method by using BM25 embeddings to cluster similar examples.
        \item Examples are selected from cluster centroids, ensuring better alignment with the input text.
        
    \end{itemize}
    
    \item \textbf{REFT\_LLAMA,REFT\_Phi3, REFT\_Qwen2}:
    \begin{itemize}
        \item These are fine-tuned versions of popular LLMs (e.g., LLAMA, Phi3, and Qwen2) using the REFT (Representation Fine-Tuning) technique.
        
    \end{itemize}
    
    \item \textbf{LLAMA3\_qlora\_small , LLAMA3\_qlora\_large}:
    \begin{itemize}
        \item These baselines involve fine-tuning LLAMA3 models using the QLoRA technique, which allows efficient fine-tuning of large language models by quantizing their weights.
        \item The \textit{small} version has fewer parameters, while the \textit{large} version delivers better performance due to its larger capacity.
    \end{itemize}
    
    \item \textbf{LLAMA3\_qlora\_large\_dpo}:
    \begin{itemize}
        \item LLAMA3 large model fine-tuned with QLoRA by applying Direct Preference Optimization (DPO).
        \item DPO aligns the model's outputs to synthetic  preferences using a preference dataset, improving metadata generation quality. The preference is selected by GPT-4o.
    \end{itemize}
    
    \item \textbf{Phi3\_qlora\_small , Phi3\_qlora\_large}:
    \begin{itemize}
        \item These are QLoRA fine-tuned versions of the Phi3 model, with small and large parameter configurations.
     \end{itemize}
    
    \item \textbf{Phi3\_qlora\_large\_dpo}:
    \begin{itemize}
        \item A fine-tuned version of Phi3 using QLoRA and Direct Preference Optimization, which significantly improves alignment and performance on text-to-metadata tasks.
        \item DPO aligns the model's outputs to synthetic  preferences using a preference dataset, improving metadata generation quality. The preference is selected by GPT-4o.
    \end{itemize}
    
    \item \textbf{Qwen\_qlora\_small , Qwen\_qlora72B\_large}:
    \begin{itemize}
        \item These baselines use the QLoRA fine-tuning method for the Qwen model.
        \item The \textit{72B large} version represents the largest Qwen model evaluated, with improved accuracy over smaller versions.
    \end{itemize}
    
    \item \textbf{Qwen\_qlora\_large\_dpo}:
    \begin{itemize}
        \item Combines QLoRA fine-tuning with DPO for the Qwen model to enhance preference-aligned outputs and metadata generation.
    \end{itemize}
    
    \item \textbf{Incontext\_learning\_with\_LLMs\_merge}:
    \begin{itemize}
        \item This baseline involves using multiple LLM agents. In this baseline, Qwen\_qlora72B\_large , Phi3\_qlora\_large\_dpo and LLAMA3\_qlora\_large\_dpo are used as LLM agents.
        \item  Here the ranker used is not a finetuned one but one that uses Incontext learning using few shot examples to choose the best metadata for the input context.
    \end{itemize}
    
    \item \textbf{Infogen (small) and Infogen (large)}:
    \begin{itemize}
        \item These configurations represent the proposed framework Infogen.
        \item The \textit{small} version uses smaller fine-tuned LLMs namely llama\_qlora\_small ,qwen\_qlora\_small and phi3\_qlora\_small, while the \textit{large} version leverages larger LLMs powered by DPO for improved accuracy and alignment.\
        \item  Here the ranker used is a fine-tuned  LLM namely LLAMA-3.1 70B model that is being trained for this task.
        
    \end{itemize}

\end{enumerate}

\subsection{More details about metrics used}
\label{sec:metrics}

The Infogen framework uses several metrics to evaluate the quality of metadata generation and infographic creation. These metrics are broadly divided into automatic metrics and human evaluation metrics. 

\textbf{Automatic Evaluation Metrics}

\textbf{1. Subchart Accuracy}
This metric measures the percentage of correctly identified sub charts in the generated metadata compared to the ground truth. It evaluates how accurately the system detects the number and types of sub charts.

\begin{scriptsize}
\textbf{
\begin{multline}
\text{Subchart Accuracy} = \\
\frac{\text{Number of Correctly Predicted Subcharts}}{\text{Total Subcharts in Ground Truth}} \times 100
\end{multline}
}
\end{scriptsize}

Where:
\begin{itemize}
    \item Correct Subcharts: Subcharts with the correct type, structure, and alignment.
    \item Total Subcharts in Ground Truth: Total subcharts present in the reference metadata.
\end{itemize}

\textbf{2. Root Square Error (RSE) for Subcharts}
This metric computes the error between the number of subcharts in the ground truth and the generated metadata.

\begin{scriptsize}
\textbf{
\begin{equation}
\text{RSE} = 
\sqrt{\frac{1}{N} \sum_{i=1}^N \left( \hat{y}_i - y_i \right)^2}
\end{equation}
}
\end{scriptsize}

Where:
\begin{itemize}
    \item \( \hat{y}_i \): Number of subcharts in the generated metadata for the \(i\)-th sample.
    \item \( y_i \): Number of subcharts in the ground truth for the \(i\)-th sample.
    \item \( N \): Total number of samples.
\end{itemize}

\textbf{3. Subchart Type Accuracy}
This metric evaluates the percentage of correctly classified subchart types in the generated metadata.

\begin{scriptsize}
\textbf{
\begin{multline}
\text{Subchart Type Accuracy} = \\
\frac{\text{Number of Correct Subchart Types}}{\text{Total Subcharts in Ground Truth}} \times 100
\end{multline}
}
\end{scriptsize}

\textbf{4. Statistical Accuracy}
This metric evaluates the accuracy of numerical values extracted from the text and included in the generated metadata. Numerical values from both the generated and ground truth metadata are sorted and compared sequentially for correctness.
\begin{scriptsize}
\textbf{
\begin{multline}
\text{Statistical Accuracy} = \\
\frac{\text{Number of Correct Data Points}}{\text{Total Data Points in Ground Truth}} \times 100
\end{multline}
}
\end{scriptsize}

\textbf{5. Textual Information Metrics}
These metrics evaluate the quality of textual components in the generated metadata.

\textbf{a. Title Rouge-L}
This metric compares the generated title with the ground truth title using the Rouge-L metric, which considers the longest common subsequence (LCS) between two texts.

\begin{scriptsize}
\textbf{
\begin{equation}
\text{ROUGE-L} = \frac{\text{LCS Length}}{\text{Reference Length}}
\end{equation}
}
\end{scriptsize}

\textbf{b. Summary Rouge-L}
This metric evaluates the quality of the summary text in the metadata, again using Rouge-L.

\begin{scriptsize}
\textbf{
\begin{equation}
\text{ROUGE-L (Summary)} = \frac{\text{LCS Length}}{\text{Reference Length}}
\end{equation}
}
\end{scriptsize}

\textbf{c. Subchart Summary Rouge-L}
The \textit{Subchart Summary ROUGE-L} metric evaluates the similarity between the summaries of sub charts in the ground truth metadata and the generated metadata. For each subchart summary in the generated metadata, the ROUGE-L score is computed against the corresponding summary in the ground truth metadata. The metric then selects the maximum ROUGE-L score among all possible matches, ensuring the highest alignment between the generated and ground truth summaries.

\begin{scriptsize}
\textbf{
\begin{multline}
\text{Subchart Summary ROUGE-L} = \\
\max_{(s_g, s_t) \in (\text{Generated Metadata} \times \text{Ground Truth Metadata})} \text{ROUGE-L}(s_g, s_t)
\end{multline}
}
\end{scriptsize}

\noindent \textbf{Explanation:}
\begin{itemize}
    \item \(s_g\): A summary from the generated metadata.
    \item \(s_t\): A summary from the ground truth metadata.
     \item \(\text{ROUGE-L}(s_g, s_t)\): The ROUGE-L score calculated between a generated summary \(s_g\) and a ground truth summary \(s_t\).
    
\end{itemize}

\textbf{Human Evaluation Metrics}
\begin{itemize}
    \item \textbf{Readability Score}: This metric ensures if  statistical charts are free from overlaps or collisions with text or barsinfographic (1–5 scale) .
    \item \textbf{Visual Score}: Evaluates the layout, design, color scheme, and accuracy in identifying subcharts in the generated infographic compared to the ground truth. The assessment is done on a scale of 1 to 5. For instance, if the ground truth contains bar charts but the generated infographic includes pie charts instead, the visual score will be penalized accordingly.
   
    \item \textbf{Data Accuracy and Alignment Score}: Measures the correctness of the data and alignment of sub-charts with respect to each other  in conveying insights, also rated from 1 to 5.
\end{itemize}

\subsection{MetaData Examples}

Below here are examples of metadata for the given infographics .\par

\textbf{Example 1:}
\begin{center}
\includegraphics[width=5cm]{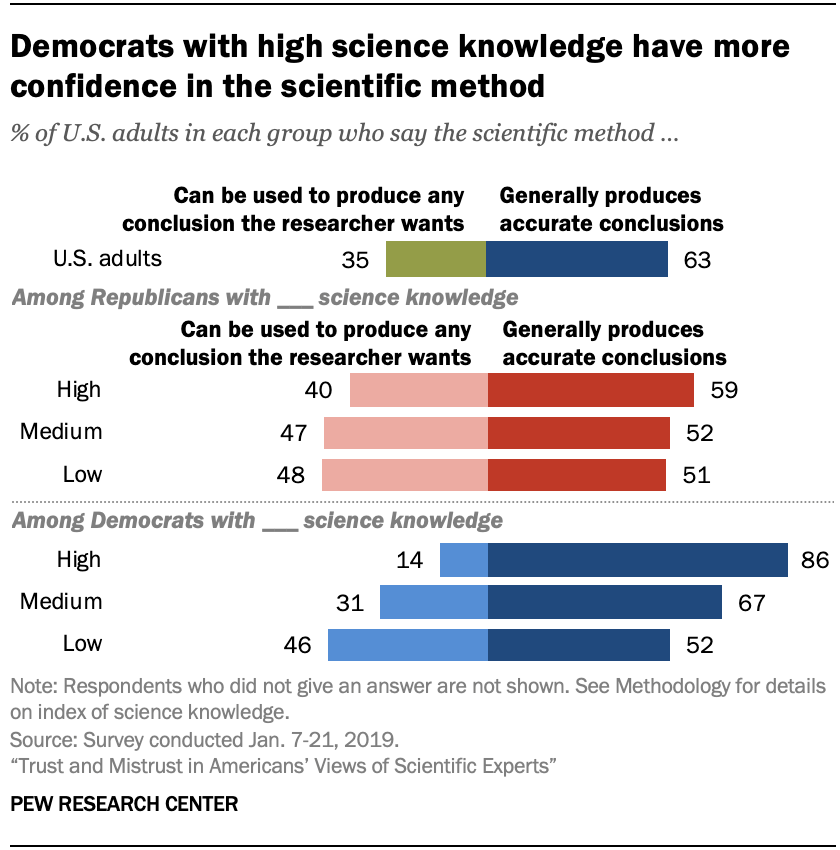}
\end{center}

\noindent \textbf{Subchart 1:} This is a bar chart. The axes are not specified for the X-axis, and the Y-axis represents U.S. adults. The statistics are: Can be used to produce any conclusion the researcher wants: 35\%, and Generally produces accurate conclusions: 63\%. The text associated with this subchart is "U.S. adults". The position of the subchart is the first one in the image, and the text is positioned above the corresponding subchart. The background is white, with dimensions of 510px width and 45px height. The fonts used are Arial, bold for the title, and Arial for the labels. \\
\textbf{Subchart 2:} This is also a bar chart. The X-axis is not specified, and the Y-axis represents science knowledge categories for Republicans (High, Medium, Low). The statistics are: High Science Knowledge: Can be used to produce any conclusion the researcher wants: 40\%, Generally produces accurate conclusions: 59\%; Medium Science Knowledge: Can be used to produce any conclusion the researcher wants: 47\%, Generally produces accurate conclusions: 52\%; Low Science Knowledge: Can be used to produce any conclusion the researcher wants: 48\%, Generally produces accurate conclusions: 51\%. The text associated with this subchart is "Among Republicans with science knowledge". The subchart is positioned below the first subchart, and the text is located above it. The background is white, with dimensions of 510px width and 50px height. The fonts used are Arial, bold for the title, and Arial for the labels. \\
\textbf{Subchart 3:} This is another bar chart. The X-axis is not specified, and the Y-axis represents science knowledge categories for Republicans (High, Medium, Low). The statistics are: High Science Knowledge: Can be used to produce any conclusion the researcher wants: 14\%, Generally produces accurate conclusions: 86\%; Medium Science Knowledge: Can be used to produce any conclusion the researcher wants: 31\%, Generally produces accurate conclusions: 67\%; Low Science Knowledge: Can be used to produce any conclusion the researcher wants: 46\%, Generally produces accurate conclusions: 52\%. The text associated with this subchart is "Among Democrats with science knowledge". The subchart is positioned below the second subchart, and the text is positioned above it. The background is white, with dimensions of 510px width and 45px height. The fonts used are Arial, bold for the title, and Arial for the labels.

\textbf{Example 2:}
\begin{center}
\includegraphics[width=5cm]{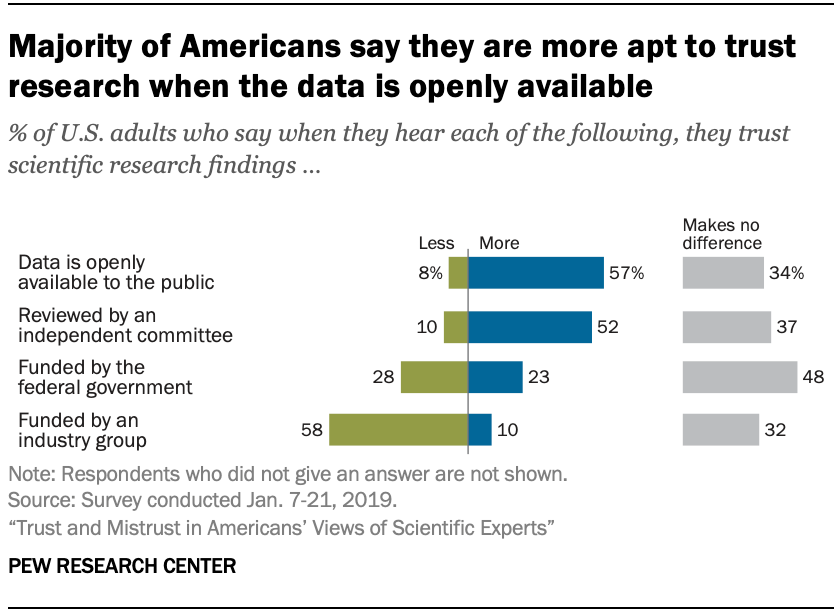}
\end{center}

\noindent \textbf{Subchart 1:} This is a horizontal bar chart. The X-axis represents percentage, and the Y-axis represents categories. The statistics are: Data is openly available to the public: Less: 8\%, More: 57\%, Makes no difference: 34\%; Reviewed by an independent committee: Less: 10\%, More: 52\%, Makes no difference: 37\%; Funded by the federal government: Less: 28\%, More: 23\%, Makes no difference: 48\%; Funded by an industry group: Less: 58\%, More: 10\%, Makes no difference: 32\%. The text associated with this subchart is "Majority of Americans say they are more apt to trust research when the data is openly available". The position of the subchart is the first chart, and the text is positioned at the top left of the corresponding subchart. The background is white, with dimensions of 612px width and 420px height. The fonts used are bold sans-serif for the title, and sans-serif for the axes and labels. \\
\textbf{Subchart 2:} This is also a horizontal bar chart. The X-axis represents percentage, and the Y-axis represents categories. The statistics are: Data is openly available to the public: Less: 8\%, More: 57\%, Makes no difference: 34\%; Reviewed by an independent committee: Less: 10\%, More: 52\%, Makes no difference: 37\%; Funded by the federal government: Less: 28\%, More: 23\%, Makes no difference: 48\%; Funded by an industry group: Less: 58\%, More: 10\%, Makes no difference: 32\%. The text associated with this subchart is "Majority of Americans say they are more apt to trust research when the data is openly available". The subchart is positioned as the second chart, to the right of the first chart, and the text is positioned similarly. The background is white, with dimensions of 612px width and 420px height. The fonts used are bold sans-serif for the title, and sans-serif for the axes and labels.

\bigskip

\textbf{Example 3:}
\begin{center}
\includegraphics[width=5cm]{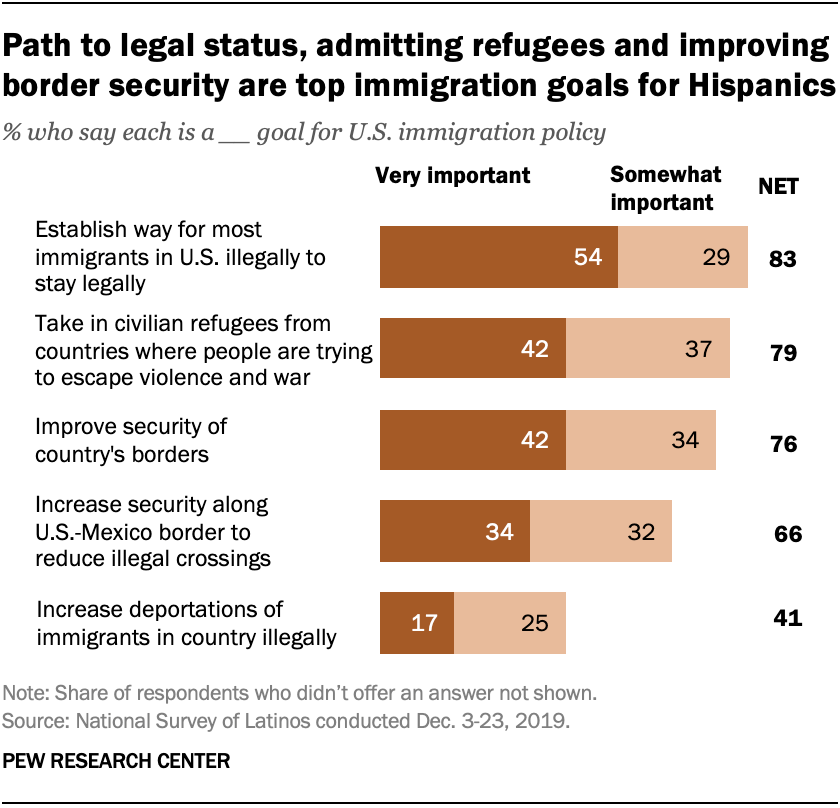}
\end{center}

\noindent \textbf{Subchart:} This is a horizontal bar chart. The X-axis represents percentage, and the Y-axis represents categories. The statistics are: Data is openly available to the public: Less: 8\%, More: 57\%, Makes no difference: 34\%; Reviewed by an independent committee: Less: 10\%, More: 52\%, Makes no difference: 37\%; Funded by the federal government: Less: 28\%, More: 23\%, Makes no difference: 48\%; Funded by an industry group: Less: 58\%, More: 10\%, Makes no difference: 32\%. The text associated with this subchart is "Majority of Americans say they are more apt to trust research when the data is openly available". The position of the subchart is the only chart, and the text is positioned at the top center of the corresponding subchart. The background is white, with dimensions of 612px width and 840px height. The fonts used are bold sans-serif for the title, sans-serif for the axes, and sans-serif for the labels.

\subsection{More Statistical Analysis}
\begin{figure}[hbt!]
    \centering
    \includegraphics[width=0.5\textwidth]{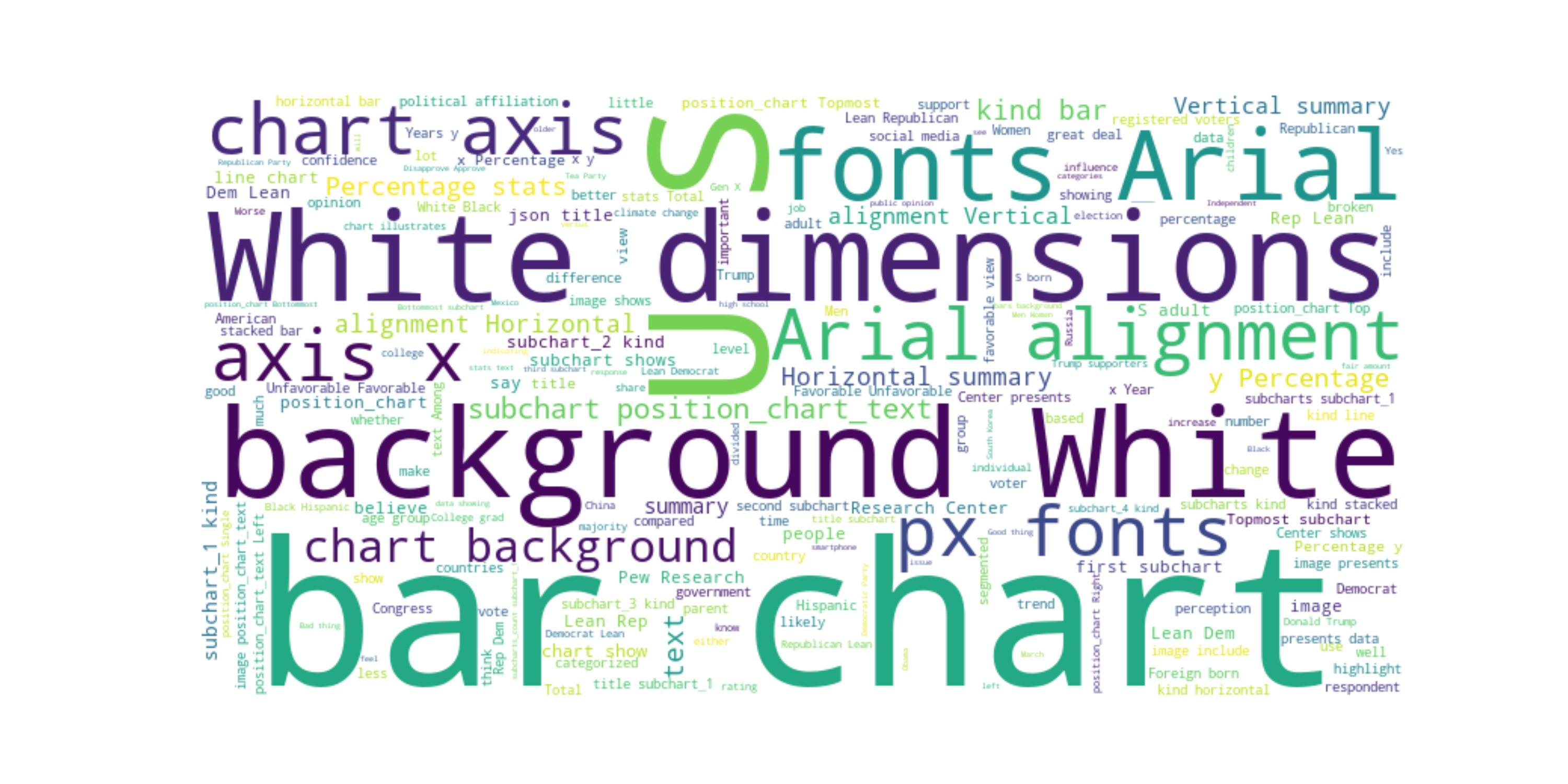}  
    \caption{Word Cloud Generated from Metadata}
    \label{fig:wordcloud}
\end{figure}

\begin{figure}[hbt!]
    \centering
    \includegraphics[width=0.5\textwidth]{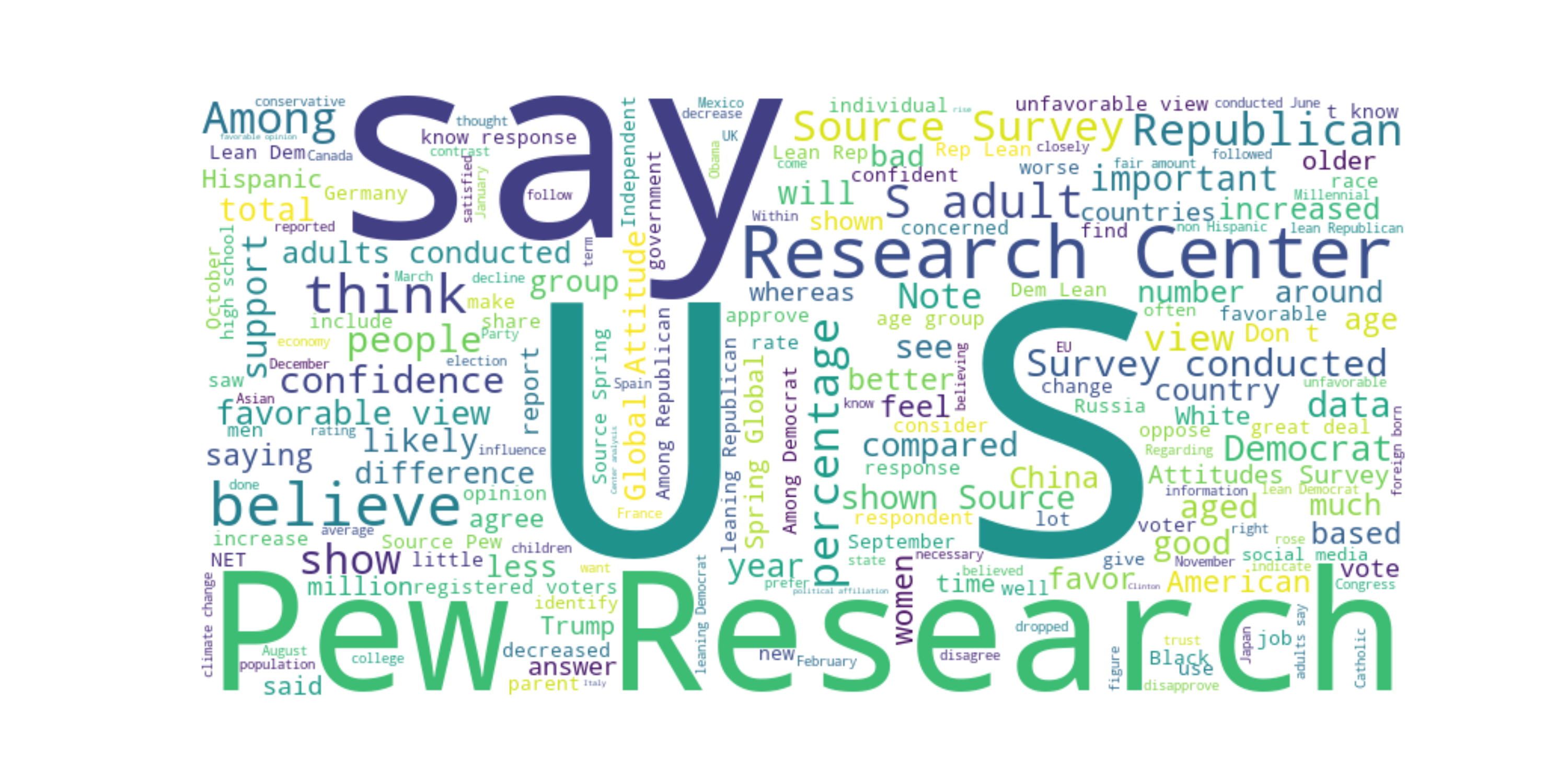}  
    \caption{Word Cloud Generated from InputText}
    \label{fig:wordcloud_text}
\end{figure}

\begin{figure}[hbt!]
    \centering
    \includegraphics[width=0.4\textwidth]{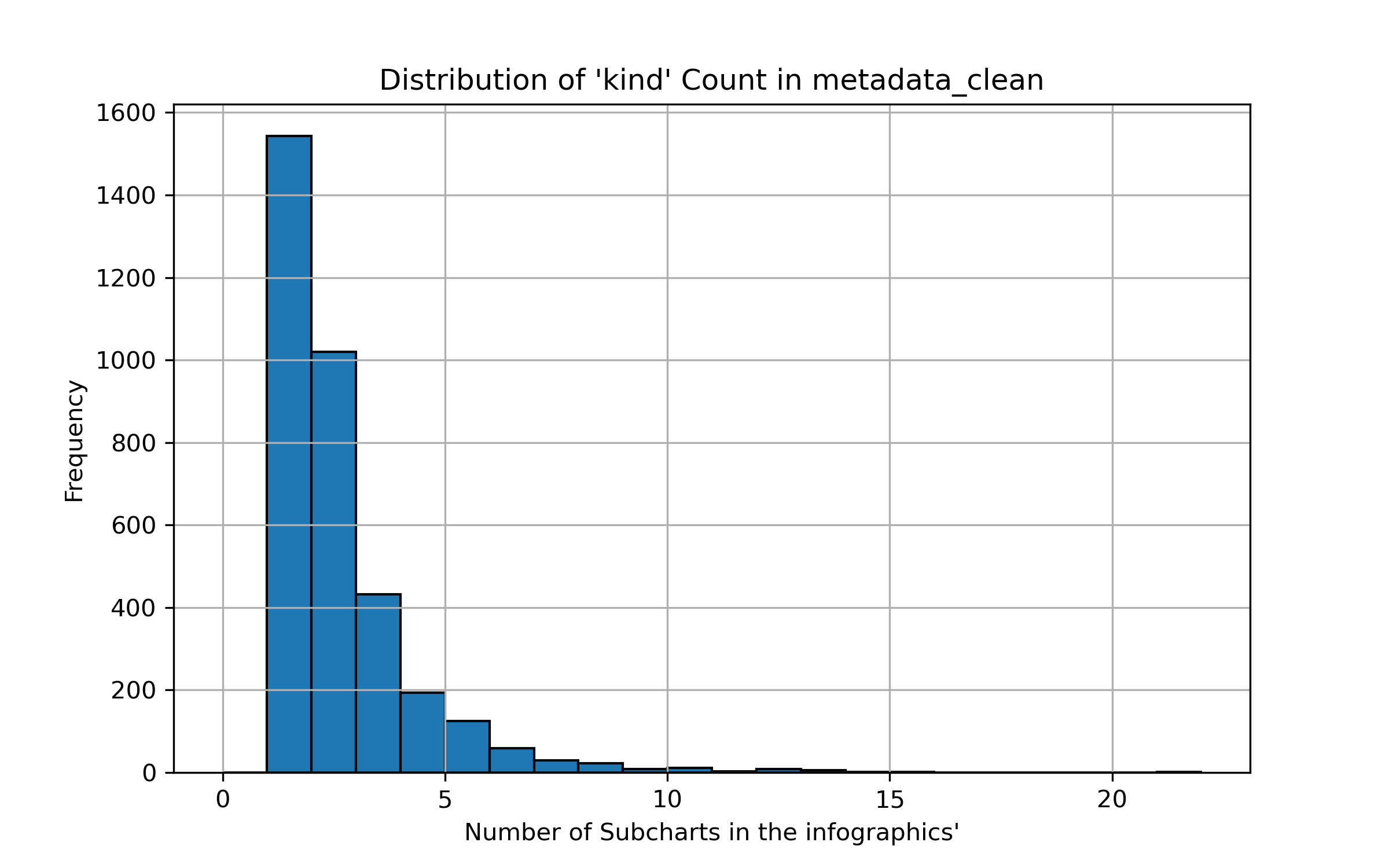}  
    \caption{Distribution of number of subcharts in \textit{\textbf{Infodat}}}
    \label{fig:kin}
\end{figure}

We also have plotted the word cloud  of metadata and input text in Figure-\ref{fig:wordcloud} and Figure-\ref{fig:wordcloud_text}. And finally we have plotted the distribution of number subcharts in the \textit{\textbf{Infodat}} dataset in Figure-\ref{fig:kin}.

\subsection{Different Prompts Used for \textit{\textbf{Infodat}} and \textit{\textbf{Infogen}} are shown below . \textit{\textbf{Infodat}} prompts are shown in blue and \textit{\textbf{Infogen}} in green . }
\begin{tcolorbox}[colback=blue!5!white, colframe=blue!75!black, title=Prompt for Detailed Chart Description]
\label{sec:prmopts}

\textit{\textbf{You are an intelligent AI assistant that can thoroughly read and understand charts. Can you give a detailed description of the attached chart in plain English, such that a human is able to reconstruct the chart based on the generated description? For each section, mention all the statistics and the associated text attached to it in the image, like the title or heading for that section. Do not mention the type of chart it is, the number of sections in the image, or use terms like 'section,' 'chart,' or 'image.' Write everything in a single passage, without sub-passages, so it feels like all information is provided in one continuous context. Ensure that for each section, all statistics and corresponding titles or text are present, and remove source information in the final output. Follow the writing style as shown in the provided examples.}}

\end{tcolorbox}

\begin{tcolorbox}[colback=blue!5!white, colframe=blue!75!black, title=Prompt used for complex infographic classification]

\textit{\textbf{You are an expert annotator with extensive knowledge of statistical charts. Your task is to evaluate whether the given infographic qualifies as a complex infographic or not. The definition of a complex infographic is as follows:}} A complex infographic contains multiple subcharts, each with an associated title, heading, and context. Additionally, it is information-dense, presenting a large amount of data in a concise manner.

\textbf{\textit{Please refer < Labelled Examples > for annotated examples. }}

\end{tcolorbox}

\begin{tcolorbox}[colback=blue!5!white, colframe=blue!75!black, title=Prompt for Detailed Subchart Analysis]

\textit{\textbf{You are an expert statistician whose task is to analyze an image containing multiple subcharts. Your goal is to provide detailed information about each subchart. Given this context, please count the total number of subcharts in the image and ensure you provide the correct number. Next, identify the type of subchart (for example, bar, line, or pie) and describe the axes by specifying their labels and units. Extract the key statistics or data points presented in the subchart. Also, identify any associated text, such as the title, heading, or other related descriptions. Determine where the subchart is positioned within the image, and describe its location relative to other subcharts. Similarly, specify where the text linked to the subchart is located. Additionally, describe the background of the subchart (whether it’s a solid color, gradient, or otherwise), and measure the dimensions of the subchart in pixels. Finally, identify the fonts used for the text in the subchart.}}

\textit{\textbf{Please provide thorough and detailed responses to each aspect. You can refer to the given examples for guidance.}}

\end{tcolorbox}

\begin{tcolorbox}[colback=green!5!white, colframe=green!75!black, title=Prompt for Metadata Selection Task]

\textit{\textbf{As a Natural Language Processing Expert, your task is to determine the superior metadata output for a given text-to-metadata generation task. You are provided with two metadata outputs, Option 1 \{metadata1\} and Option 2 \{metadata2\}, both generated by the same model but with different temperature settings. Your goal is to select the metadata that most accurately represents the information derived from the input text. Make sure the selected metadata captures all relevant details and aligns with the intended structure of the task. The output should consist solely of the selected option, without any additional information beyond the provided choices.}}

\end{tcolorbox}

\begin{tcolorbox}[colback=green!5!white, colframe=green!75!black, title=Prompt for Ranker LLM]

You are a statistician expert tasked with selecting the best metadata for the Input\_text\_clean: [insert text here]. Your goal is to choose the most optimal metadata based on the following questions:

1. Title of the context? 
2. Summary of the context (what is being inferred)? 
3. How many subcharts are there? For each subchart, identify: 
   - Type of subchart 
   - Axes (including associated text) 
   - Statistics 
   - Text (title or heading) 
   - Position relative to other subcharts 
   - Position of text 
   - Background type 
   - Dimensions (in px) 
   - Fonts (one-word description) 
   - Alignment (horizontal or vertical) 
   - Subchart summary (include any text that doesn’t fit the title or heading).

Return the response in JSON format with this structure:
{
    "title": {},
    "summary": {},
    "subchart\_1": {
        "kind": {},
        "axis": {},
        "stats": {},
        "text": {},
        "position\_chart": {},
        "position\_chart\_text": {},
        "background": {},
        "dimensions": {},
        "fonts": {},
        "alignment": {},
        "summary": {}
    },
    "subchart\_2": { ... }
}

You will be given three options. Choose the most optimal metadata, focusing first on the correct number of subcharts. The metadata should align properly, with correct textual insights and statistics. Provide only the best option as output.

\end{tcolorbox}

\begin{tcolorbox}[colback=green!5!white, colframe=green!75!black, title=Prompt for Coder LLM, width=\textwidth/2]

You are an expert coder with strong graphic understanding. You will be provided with metadata to generate Python code for a final infographic. The metadata answers:

- Title of the context?
- Summary of the context (what is it inferring)?
- Number of subcharts? For each subchart:
  1) Subchart type
  2) Axes (with text)
  3) Stats of the subchart
  4) Text (title or heading)
  5) Subchart position relative to others
  6) Position of text
  7) Background type
  8) Dimensions (px)
  9) Fonts (one-word)
  10) Alignment (horizontal or vertical)
  11) Subchart summary (include text not in title or heading)

Return the response in JSON format:
\{ "title": \{\}, "summary": \{\}, "subchart\_1": \{ "kind": \{\}, "axis": \{\}, "stats": \{\}, "text": \{\}, "position\_chart": \{\}, "background": \{\}, "dimensions": \{\}, "fonts": \{\}, "alignment": \{\}, "summary": \{\} \}, "subchart\_2": \{...\} \}.

Key points for Python code:
1) Ensure the number of subcharts matches the metadata. For bar charts, bars should be on the y-axis and sized properly.
2) Maintain correct axes, stats, and chart positions. The chart should be visually appealing, with no extra axes.
3) Include title and summary, ensure alignment, and avoid cramped spacing.

Tips:
- Adjust vertical spacing to avoid crowding.
- Use distinct colors, avoiding grey/black.
- Place data values inside bars for readability.
- Center the title and use a clean layout.
- Adjust height for proper subchart fitting.

Use Python libraries like Plotly or Plotnine. The output should only contain Python code, no JSON or additional information.

\end{tcolorbox}

\begin{tcolorbox}[colback=green!5!white, colframe=green!75!black, title=Judge LLM Prompt, width=\textwidth/2]

You are a judge LLM with expertise in evaluating Python code and graphical visualizations. Your task is to assess the provided Python code against the metadata. Based on the metadata, determine whether the code satisfies all the following constraints:

- **Subchart Count**: Ensure the code generates exactly the number of subcharts specified in the metadata.
- **Subchart Type**: Verify that each subchart (e.g., bar chart, line chart) matches the specified type.
- **Axes and Stats**: Ensure the axes (including text) and statistics of each subchart match the metadata.
- **Subchart Position**: Check that the position of each subchart and its associated text is correct relative to other subcharts.
- **Alignment and Layout**: Confirm that the alignment (horizontal or vertical) and layout of subcharts follow the metadata and are visually appealing.
- **Fonts and Dimensions**: Make sure the fonts and dimensions (in px) for each subchart match the metadata.
- **Background and Colors**: Verify that the background type and colors (avoid grey/black) are correct and distinguish between elements.
- **Title and Summary**: Ensure the overall title and summary from the metadata are reflected in the final output.

Additionally, check for the following:
- Vertical spacing should avoid cramped visuals.
- Data values should be visible inside bars for bar charts.
- No extra axes should be added beyond what’s described in the metadata.
- There should be no errors such as `ValueError: Vertical spacing cannot be greater than (1 / (rows - 1))`.

If you are statisfied with the code replly with "Yes" . Else your feedback should specify whether all the constraints are met, and if not, provide specific details on what is missing or incorrect in the code.

\end{tcolorbox}

\end{document}